\documentclass[journal,twoside,web]{ieeecolor}
\usepackage{generic}
\usepackage{cite}
\usepackage{amsmath,amssymb,amsfonts}
\usepackage{algorithmic}
\usepackage{graphicx}
\usepackage{algorithm,algorithmic}
\usepackage{hyperref}
\hypersetup{hidelinks}
\usepackage{textcomp}
\usepackage{multirow}
\usepackage{makecell} 
\usepackage{booktabs}
\usepackage{threeparttable}
\def\BibTeX{{\rm B\kern-.05em{\sc i\kern-.025em b}\kern-.08em
    Tn\kern-.1667em\lower.7ex\hbox{E}\kern-.125emX}}
\usepackage{verbatim}
\usepackage[utf8]{inputenc}
\begin{document}
\title{A Patient-Specific Pulmonary Arterial Tree Digital Twin to Extract Pulmonary Embolism Biomarkers}

\author{Morgane des Ligneris, Nathan Painchaud, Allan Serva, Laurent Bertoletti, Pierre Croisille, Carole Frindel, Odyssée Merveille
\thanks{XX/XX/25 Date of submission. This work was funded by the French ANR through the PERSEVERE and PreSPIN projects (ANR-22-CE45-0018, ANR-20-CE45-0011) and within the framework of the LABEX PRIMES (ANR-11-LABX-0063). This work was also performed using HPC resources from GENCI-IDRIS (Grant 2025-AD011014039R2).}
\thanks{M. des Ligneris is with Univ Lyon, INSA‐Lyon, Université Claude Bernard Lyon 1, UJM-Saint Etienne, CNRS, Inserm, CREATIS UMR 5220, U1294, F‐69621, Lyon, France}
\thanks{N. Painchaud is with Univ Lyon, INSA‐Lyon, Université Claude Bernard Lyon 1, UJM-Saint Etienne, CNRS, Inserm, CREATIS UMR 5220, U1294, F‐69621, Lyon, France}
\thanks{A. Serva is with the Department of Pneumology, CHU Saint-Etienne, UJM Saint-Etienne, France and Université Jean Monnet Saint-Étienne, CHU Saint-Étienne, Mines Saint-Étienne, INSERM, SAINBIOSE U1059, CIC 1408, Département de Médecine Vasculaire et Thérapeutique, F-CRIN INNOVTE network, all in F-42055, Saint-Étienne, France}
\thanks{Laurent Bertoletti is with Université Jean Monnet Saint-Étienne, CHU Saint-Étienne, Mines Saint-Étienne, INSERM, SAINBIOSE U1059, CIC 1408, Département de Médecine Vasculaire et Thérapeutique, F-CRIN INNOVTE network, all in F-42055, Saint-Étienne, France}
\thanks{Pierre Croisille is with Univ Lyon, INSA‐Lyon, Université Claude Bernard Lyon 1, UJM-Saint Etienne, CNRS, Inserm, CREATIS UMR 5220, U1294, F‐69621, Lyon, France and with Department of Radiology, CHU Saint-Etienne, UJM Saint-Etienne, Saint-Etienne, France.}
\thanks{C. Frindel is with IUF, Institut Universitaire de France, Paris and with Univ Lyon, INSA‐Lyon, Université Claude Bernard Lyon 1, UJM-Saint Etienne, CNRS, Inserm, CREATIS UMR 5220, U1294, Lyon, France (e-mail: carole.frindel@creatis.insa-lyon.fr).}
\thanks{O. Merveille is with Univ Lyon, INSA‐Lyon, Université Claude Bernard Lyon 1, UJM-Saint Etienne, CNRS, Inserm, CREATIS UMR 5220, U1294, F‐69621, Lyon, France (e-mail: odyssee.merveille@creatis.insa-lyon.fr).}}

\maketitle

\begin{abstract}
Pulmonary embolism, the obstruction of a pulmonary artery by a blood clot, is one of the leading causes of acute cardiovascular syndrome. In clinical practice, therapeutic decisions after diagnosis via computed tomography pulmonary angiography rely on risk stratification, which categorizes 30-day mortality risk into three categories. This stratification depends on the right-to-left ventricular diameter ratio and blood levels of two cardiac enzymes. However, blood biomarkers are not always available in emergency settings, and manual calculation of established severity scores -- such as Qanadli and Mastora -- is time-consuming and rarely performed in clinical routine practice. 
This study introduces an automated pipeline that models a directed graph representation of the pulmonary arterial tree, labeling its hierarchical structure and characterizing pulmonary embolism. The pipeline derives image-based biomarkers, including local artery-level features (morphological information, hierarchical position, clot volume, and resulting obstruction) and global patient-level biomarkers such as automatically calculated severity scores (Qanadli and Mastora) and the total embolic volume distribution by lobes and hierarchical levels. Using artificial-intelligence-generated binary masks of arteries, emboli, lungs, and lobes, it creates a patient digital twin of the arterial structure.
Validation of the pipeline through comparison to an existing pipeline, anatomical expectations, and manual severity score calculations demonstrates the pipeline's ability to automatically generate anatomically accurate digital twins and severity scores with strong agreement.  This supports the potential of these image-derived biomarkers to automatically provide rapid, precise information on thrombotic burden and spatial clot distribution.  
\end{abstract}


\section{Introduction}
\label{sec:introduction}
\IEEEPARstart{P}{ulmonary}
embolism (PE) is defined as the obstruction of a pulmonary artery by a blood clot, also called a thrombus. It is the third most frequent acute cardiovascular syndrome, after myocardial infarction and stroke \cite{Konstantinides2020}. In patients with PE, death may result either from cardiac repercussions or from underlying comorbidities. Incidence varies across countries, ranging from 39 cases per 100,000 in Hong Kong to 115 cases per 100,000 in the United States \cite{wendelboe2016}. 
Computed Tomography Pulmonary Angiography (CTPA) is now the primary imaging modality for PE diagnosis \cite{mehdipoor2020}, with its use increasing from 47\% to 90\% of patients between 2001 and 2013 \cite{jimenez2016}.
Clinical management depends on the severity of PE, assessed through risk stratification into low, intermediate, or high 30-day mortality categories. This stratification guides the treatment choice, from anticoagulants alone to more invasive procedures such as thrombectomy, and helps determine whether patients should be discharged, treated on the ward or admitted to intensive care.

% Guidelines
Since 2019, European guidelines \cite{Konstantinides2020} recommend that, after assessing hemodynamic instability, risk stratification should rely on three biomarkers: the right-to-left ventricular diameter ratio (RV/LV), and the blood levels of B-type natriuretic peptide (BNP) and troponin. Although the guidelines specify that the RV/LV ratio should ideally be measured manually using a transthoracic echocardiography (TTE), this exam is not routinely performed in emergency settings. In practice, the ratio is therefore measured directly from the diagnostic CTPA. An RV/LV ratio greater than one indicates dilation of the right ventricle, reflecting increased pressure and acute strain, which may result from PE. 
Elevated BNP and troponin levels are associated with cardiac stress, including myocardial stretch or right ventricular ischemia caused by pressure overload.
In clinical routine, the RV/LV is available as it is measured on the diagnostic CTPA. However, the blood test required to assess the BNP and troponin levels is not consistently performed, particularly in emergency settings, and as a result, risk stratification is often incomplete \cite{naum2024, liu2021}.
In such situations, clinicians rely on their experience, available data (age, comorbidities, blood pressure...) and the CTPA. Yet, they lack a comprehensive view of the pulmonary arterial tree (PAT), for instance, the overall degree of obstruction or the relative positions of multiple thrombi—information that could improve their understanding of the patient’s condition and support more informed treatment decisions. Severity scores to characterize the thrombotic burden exist, but the manual calculation is too complex and time-consuming so it is rarely performed in clinical practice \cite{sun2020a, djahnine2024}.

In this work, we address the impracticability of manually calculating established severity scores, and the lack of a comprehensive view of the PAT.
Our contributions are as follows : 
\begin{itemize}
    \item We designed an automatic pipeline modeling the pulmonary arterial tree (PAT) of PE patients as a directed graph directly derived from CTPA.
    \item We propose several morphological biomarkers, either local ones capturing vascular and embolic characteristics, or patient-level ones providing global indications on the arterial environment.
    \item We develop algorithms to extract these biomarkers automatically from the PAT graph and the CTPA, resulting in a directed graph-based digital twin of PE patient's arterial system and severity scores.
\end{itemize}

\section{State of the art} 

\subsection{Thrombotic burden} 
The thrombotic burden is the estimation of the obstruction of the thrombus and its impact, helping to characterize the severity. The thrombotic load can be defined in several ways, and over the decades, several scoring systems have been developed to standardize and improve the assessment of this burden, each offering unique advantages and addressing the limitations of its predecessors. 
The location within the hierarchy is important for interpreting embolism severity \cite{vedovati2012, vedovati2013, cimsit2015, paez-carpio2025}; you can observe the PAT hierarchical organization (Fig.\ \ref{fig:hier_level}). A clot is defined as proximal if located in the main pulmonary artery (MPA) emerging from the right ventricle (RV), dividing into the right or left pulmonary artery (RPA or LPA), and then the lobar arteries (levels of depth 1, 2, or 3, respectively). From the segmental arteries (level of depth 4) and onward, including the sub-segmental arteries (level 5) and those below (level 6+), a clot is considered distal. The proximal/distal distinction helps assess the proximity of the obstruction to the heart and the extent of the affected vascular territory. Hierarchical localization is therefore a key criterion. The different scores often refer to the segmental arteries. 

\subsection{Literature scores}
% Qanadli score
The Qanadli score \cite{Qanadli2001} combines the segmental focus of the Miller score \cite{Miller1971} with the Walsh score’s distinction between partial and complete obstructions \cite{Walsh1973}. For each PE, it considers the number of distal segments emerging from the obstruction, multiplied by a weight reflecting the degree of obstruction: 0 for none, 1 for partial, and 2 for complete. Isolated sub-segmental PEs are uniformly assigned 1 point. The goal was to improve the reproducibility, since the previous scores had high inter-observer variability and low sensitivity to partial obstruction.  
The Mastora score \cite{Mastora2003} refines this approach by quantifying the obstruction as a continuous percentage divided into a five-level grading system (1: 0–24\%, 2: 25–49\%, 3: 50–74\%, 4: 75–99\%, 5: 100\%). It provides higher precision at the cost of increasing complexity in calculation.
Despite their clinical relevance, the manual and time-consuming nature of these scores restricts their use in clinical routine. Automating the computation of these scores could ease their adoption and provide a more reproducible assessment of PE severity \cite{li2017, sun2020a, meyer2021, djahnine2024, xi2024}. 
The Total Embolic Volume (TEV) \cite{Huang2022} biomarker measures the total pixel volume of PE visible on CTPA, using a semi‑automatic 3D segmentation method. It has been shown to correlate with various clinical parameters, including the RV/LV ratio on CT, right ventricular dysfunction on echocardiography, and shock. Notably, TEV demonstrates a strong correlation with the Qanadli score when the TEV is less than 10 cm³, but tends to diverge for larger PE volumes. 
But TEV remains dependent on accurate clot segmentation, a task that is still often performed manually or semi-automatically, requiring substantial time and operator input \cite{furlan2011, aydin2023, pu2023a}.
Nevertheless, an automated calculation of TEV repartition by lobes or hierarchical level could complement the severity score by capturing the proximal vs. distal nature of the clots. 

Because CTPA is systematically performed in PE diagnosis, the ambition is to extract new biomarkers from the CTPA that could complement of substitute for sometimes missing blood in emergency settings in clinical practice. These new biomarkers would fall into two categories. \textit{Local biomarkers} that capture vascular and embolic features and \textit{patient-level biomarkers} that aggregate these local measurements into global indicators, such as TEV repartition by lobes and levels, and established state-of-the-art severity scores (Qanadli, Mastora) that are otherwise too complex to compute in routine clinical practice.

\subsection{PAT modeling} 
To extract novel imaging-based biomarkers directly from CTPA scans, we propose a directed graph-based representation of the pulmonary arterial tree (PAT). This approach requires two essential inputs: a segmentation mask of the PAT and a segmentation mask of the blood clot (embolus). The graph model aims to capture the hierarchical and morphological properties of the vascular structure, enabling the extraction of clinically relevant features at both local (artery-level) and global (patient-level) scales.
A critical prerequisite for PAT modeling is the accurate segmentation of the pulmonary arteries. However, current segmentation methods do not guarantee vascular connectivity \cite{keshwani2020, carneiro-esteves2025}. The absence of a fully connected vascular network can disrupt downstream processes, such as centerline extraction and graph construction.
The extraction of the vascular centerline, or skeletonization, allows for simplifying and preserving the topological information of the PAT. This process iteratively thins the segmented volume while maintaining its connectivity, ultimately yielding a 1-voxel-thick representation of the vascular network \cite{Lee1994}. The resulting skeleton serves as the backbone for graph construction, as it captures the branching structure and spatial relationships of the arteries.
The conversion of the centerline into a graph structure allows for the representation of the PAT as a network of nodes (bifurcations or terminations) and edges (artery segments). Each edge can be annotated with geometric attributes, such as length, radius, and spatial coordinates, enabling quantitative analysis of the vascular architecture.
Existing pipelines, such as \textit{VesselGraph} \cite{drees2021} from the Voreen framework \cite{meyer-spradow2009}, provide an example for centerline extraction and graph modeling. In this study, we compare our method with the VesselGraph pipeline to evaluate our graph construction process.

\section{Method}
\begin{figure*}[htbp!]
\centering
\includegraphics[width=\textwidth]{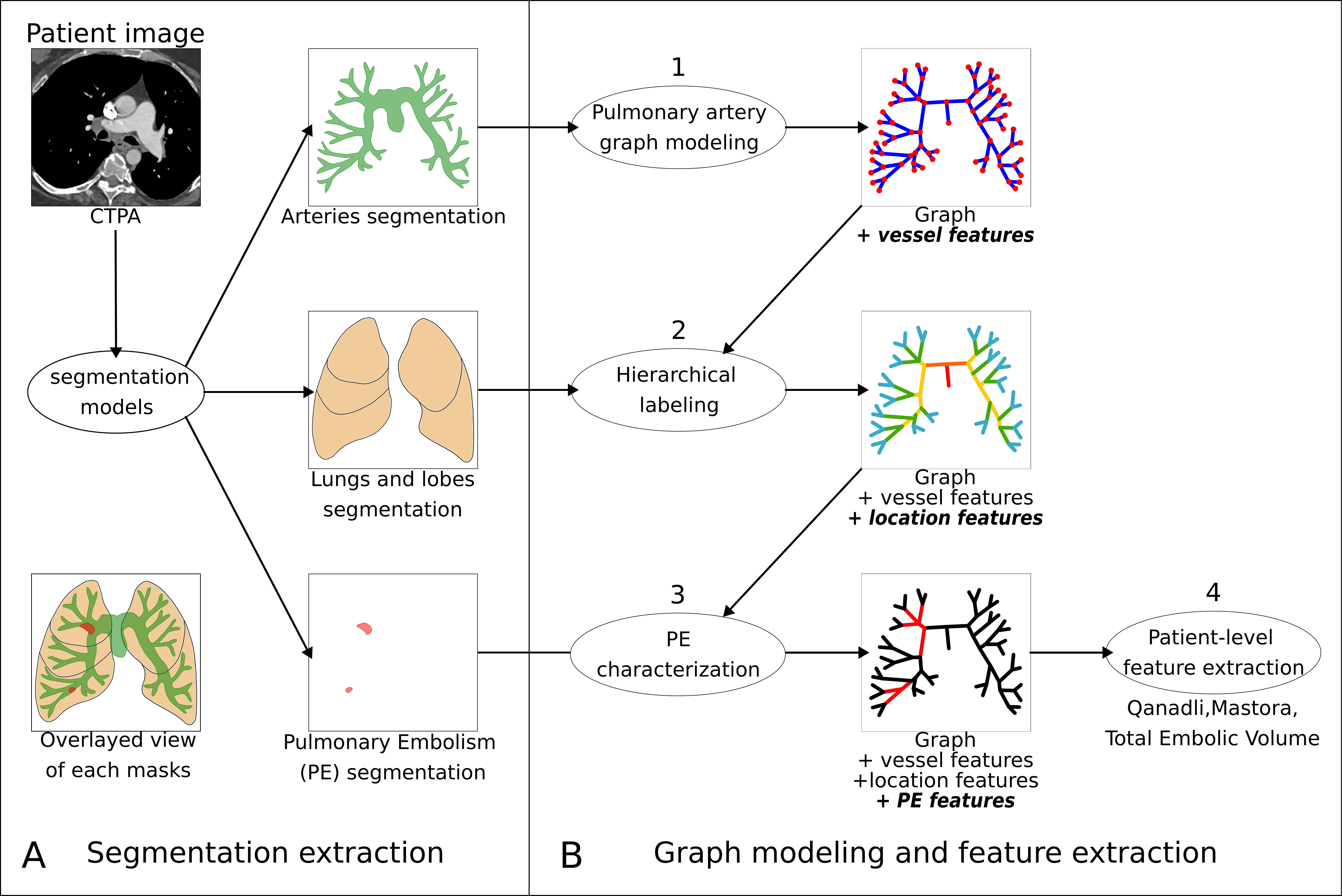}
\caption{General workflow of the proposed pipeline.
A) Input data: automatically segmented masks of arteries, PE, lungs, and lobes. B) Pipeline steps: (1) graph modeling from the arterial mask, (2) hierarchical labeling using lung and lobe masks, and (3) integration of obstruction-related information from the PE masks. Each step produces local biomarkers, which are subsequently aggregated in (4) patient-level biomarkers and global scores.}
\label{fig:general_pipeline}
\end{figure*}

The objective of the proposed pipeline is to model the PAT as a directed graph from CTPA-derived segmentations and to automatically extract morphological features, calculate established severity scores, and spatial embolic distribution. As illustrated in Fig.\ \ref{fig:general_pipeline}, the pipeline takes as input several anatomical and pathological masks. From the arterial mask, a directed graph representation of the PAT is constructed. The lung and lobe masks are then used to infer the hierarchical and anatomical localization of each arterial branch starting from the root corresponding to the main pulmonary artery (MPA). Finally, the PE mask is used to compute obstruction-related information and associate it with the corresponding graph branches. Each step of the pipeline is detailed in Section \ref{subsec:graph_modelling}. Local graph features are subsequently aggregated to derive patient-level features, and the biomarker extraction process is described in Section \ref{subsec:biomarkers_extraction}.

\subsection{Graph modeling}
\label{subsec:graph_modelling}
The graph modeling step comprises three main stages: graph construction, hierarchical labeling, and PE characterization.

\subsubsection{Pulmonary artery graph modeling}.
This step constructs a directed graph representation of the pulmonary arterial tree from the arterial segmentation mask. The process includes preprocessing, graph construction, artifact correction, and orientation of the vascular tree.

\paragraph{Preprocessing}
The preprocessing step converts the arterial binary mask into two representations: a topological skeleton and a distance map. The skeleton is obtained using the iterative thinning algorithm of Lee et al.\cite{Lee1994} from the \textit{skeletonize} function from the library \textit{skimage.morphology}, which progressively removes surface voxels from a 3x3x3 neighborhood of a pixel, while preserving the original network topology, resulting in a 1-voxel-thick structure that accurately captures the PAT architecture. In parallel, an Euclidean distance transform is applied to the arteries mask so that each artery voxel expresses the shortest distance to the artery boundary, which is computed in physical space using the voxel spacing from the CTPA metadata, allowing for the radius estimation later on. Skeletonization is sensitive to noise or surface irregularities, and can create spurious branches and cycles. These artifacts are addressed in the correction graph step. 

\paragraph{Graph construction}

\begin{figure}[htbp!]
\centering
\includegraphics[width=\columnwidth]{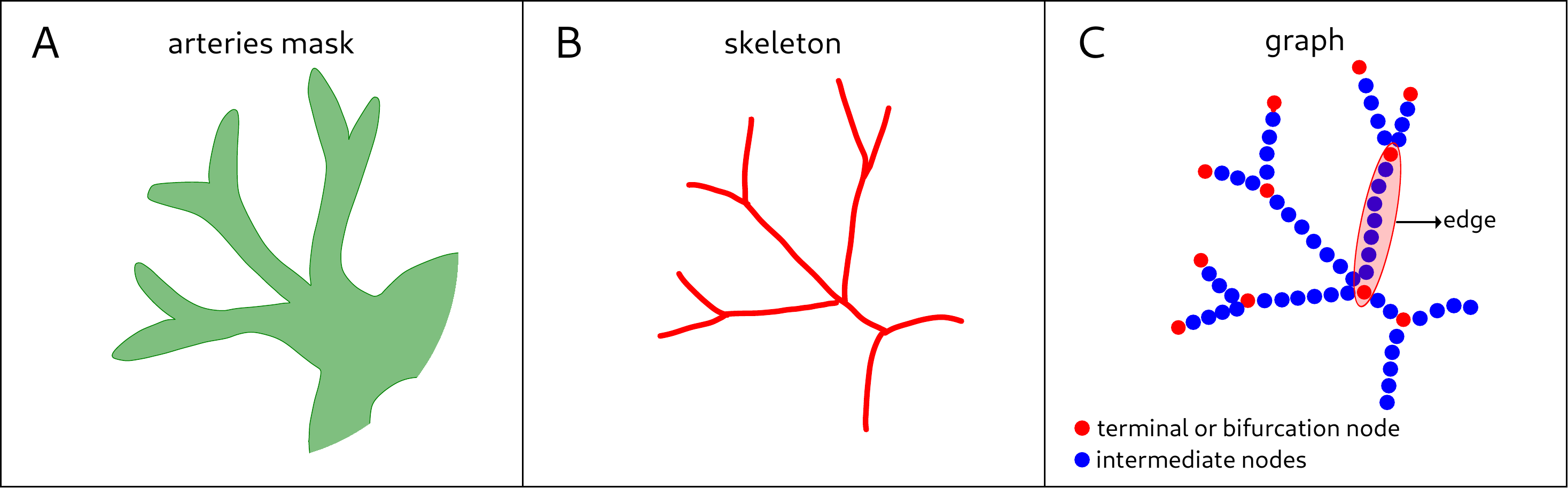}
\caption{Graph construction steps. A) Arterial mask B) Skeleton (centerline) of the arterial mask C) Voxel-level graph in which neighboring centerline voxels are connected. Bifurcations and terminal voxels are identified as nodes, and paths between them are merged to form edges of the branch-level graph.}
\label{fig:graph_construction}
\end{figure}

The graph is constructed by transforming the skeleton into a branch-level representation (Fig.\ \ref{fig:graph_construction}). First, the 3D skeleton is converted into a voxel-level graph using the \texttt{skan} library \cite{nunez-iglesias2018}, where each skeleton voxel is represented as a node connected to its neighboring voxels according to spatial adjacency. Each voxel-node is assigned its spatial 3D coordinates and a local radius value obtained from the distance map. The voxel-level graph is then simplified into a branch-level graph by classifying nodes according to their degree : terminal nodes (degree 1), bifurcation nodes (degree $\ge 3$), and intermediate nodes (degree 2). Linear chains of intermediate nodes connecting two terminal or bifurcation nodes are merged into a single edge, representing one arterial segment. The resulting undirected graph represents arteries as edges characterized by their centerline coordinates, radius profile, and length. Skeletonization-induced artifacts are handled in the subsequent graph correction step. 

\paragraph{Graph correction}

\begin{figure}[!t]
\centering
\includegraphics[width=\columnwidth]{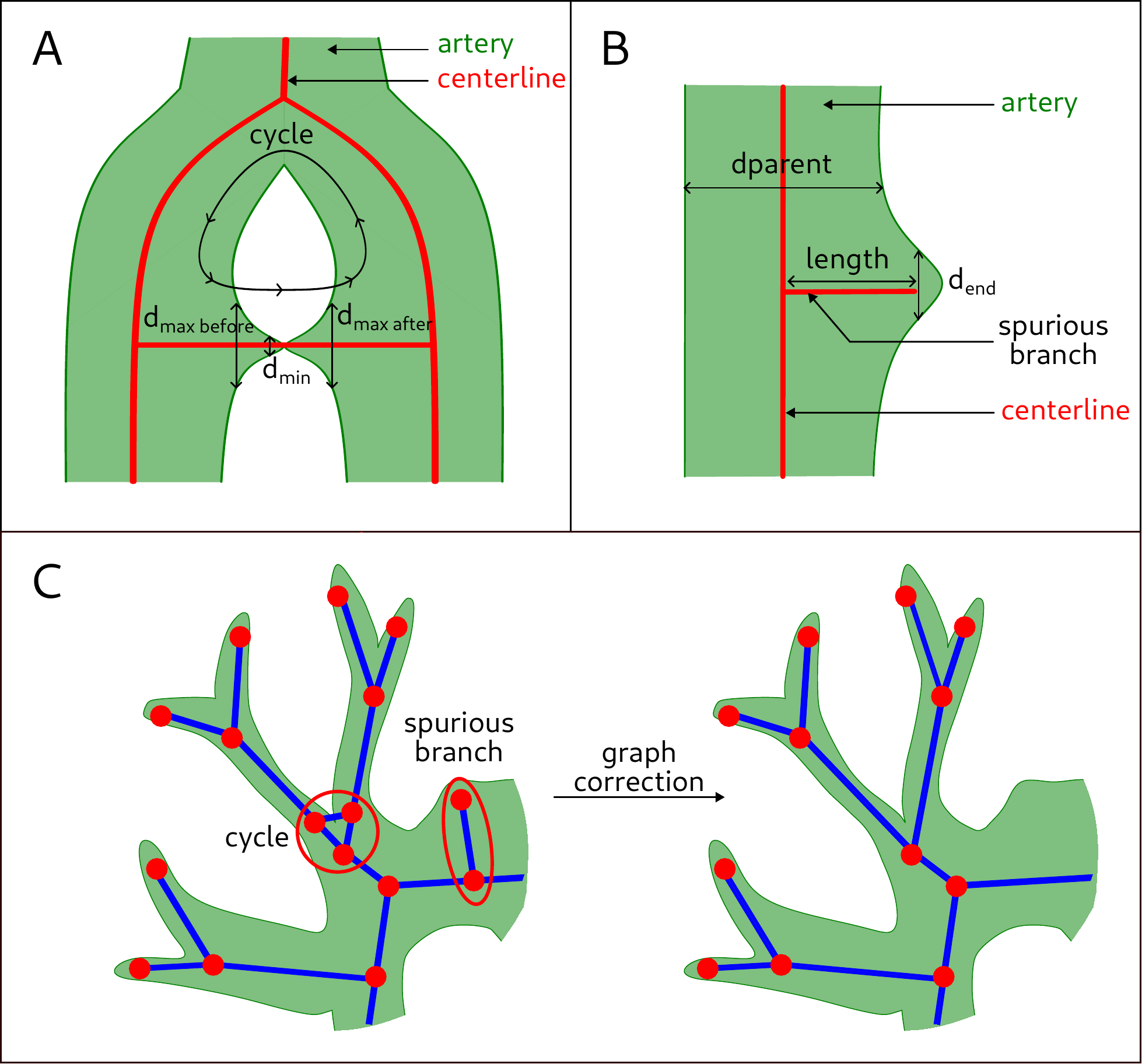}
\caption{Graph correction steps. A) Non-physiological cycle caused by a contact between two adjacent arteries in the segmentation mask, producing an artificial local diameter constriction along a single artery segment, characterize by successive diameters $d_{\text{max\ before}}, d_{\text{min}}, d_{\text{max\ after}}$. B) Spurious terminal branch generated by surface irregularities of the segmentation mask. $d_{\text{parent}}$ denotes the diameter of the parent artery and $d_{\text{end}}$ the diameter at the end of the spurious branch. C) Detection criteria used for graph correction. A cycle is detected from a constriction pattern when both the diameter decrease and subsequent increase ratios exceed a fixed threshold. A spurious branch is detected when its length is short relative to the parent diameter and its terminal diameter is strongly reduced.}
\label{fig:graph_correction}
\end{figure}

The graph correction addresses two main types of artifacts introduced during skeletonization: (i), non-physiological cycles caused by contacts between adjacent arteries (Fig.\ \ref{fig:graph_correction}A), and (ii) terminal spurious branches generated by surface irregularities of the segmentation mask (Fig.\ \ref{fig:graph_correction}B). 

For each detected cycle, we analyze the diameter profile along the artery centerline of each edge involved to identify characteristic constriction patterns. Such pattern are indicative of a false connection pinching between two distinct arteries.
For each edge belonging to detected cycle, we first identify the minimum diameter along the centerline, denoted $d_{\min}$, as well as the local maximum immediately before and after this minimum, denoted $d_{\text{max before}}$ and after $d_{\text{max after}}$, respectively. The severity of the constriction is quantified using two dimensionless ratios. The decrease ratio is defined as: 
\begin{equation}
r_{dec} = \frac{d_{max\ before} - d_{\min}}{d_{max\ before}},
\end{equation}
and the increase ratio as: 
\begin{equation}
r_{inc} = \frac{d_{max\ after} - d_{\min}}{d_{max\ after}} .
\end{equation}
An edge is removed if both ratios satisfy $r_{dec} \geqslant \tau$ and $r_{inc}\geqslant \tau$, with $\tau = 0.5$.

Spurious terminal branches are expected to be both short and to exhibit a strong distal diameter reduction. They are therefore characterized using their length $L$ and their terminal diameter $d_\text{end}$, relative to the diameter of the parent artery $d_\text{parent}$. Branch length is evaluated using the criterion:
\begin{equation}
L < \alpha*d_\text{parent}, 
\end{equation}
with $\alpha = 1.5$, ensuring that the branch is predominantly contained within the parent artery region. The distal diameter reduction is quantified by the ratio :
\begin{equation}
r_{end} = \frac{d_\text{end}}{d_\text{parent}},
\end{equation}
and a branch is considered spurious if $r_{end} < \beta$, with $\beta = 0.3$. A diameter reduction of more than 70\%, combined with a length short relative to the parent diameter, qualifies the branch as an artifact. After deletion, nodes with degree 2 are merged by concatenating centerline coordinates and radius, and by summing edge lengths. To avoid removing major arteries, the terminal artery with the largest mean diameter is excluded from deletion. 

\paragraph{Graph orientation}
Graph orientation is performed to establish a consistent proximal--distal ordering of the PAT. This ordering is required for hierarchical labeling and for any subsequent analysis relying on propagation of attributes along the vascular tree. The anatomical root of the graph is defined as the terminal artery with the largest diameter, corresponding to the main pulmonary artery (MPA). Starting from this root, edge directions are propagated using a breadth-first search (BFS) algorithm. Each edge visited is oriented from the root to the terminal branches and its geometrical attributes (including centerline coordinates and radius profiles) are reordered accordingly to ensure consistency with the assigned direction. This procedure results in a cleaned and fully oriented patient-specific pulmonary arterial tree (PAT) model. 

\subsubsection{Hierarchical labeling}
\label{subsec:hier_labeling}

\begin{figure}[htbp!]
\centering
\includegraphics[width=\columnwidth]{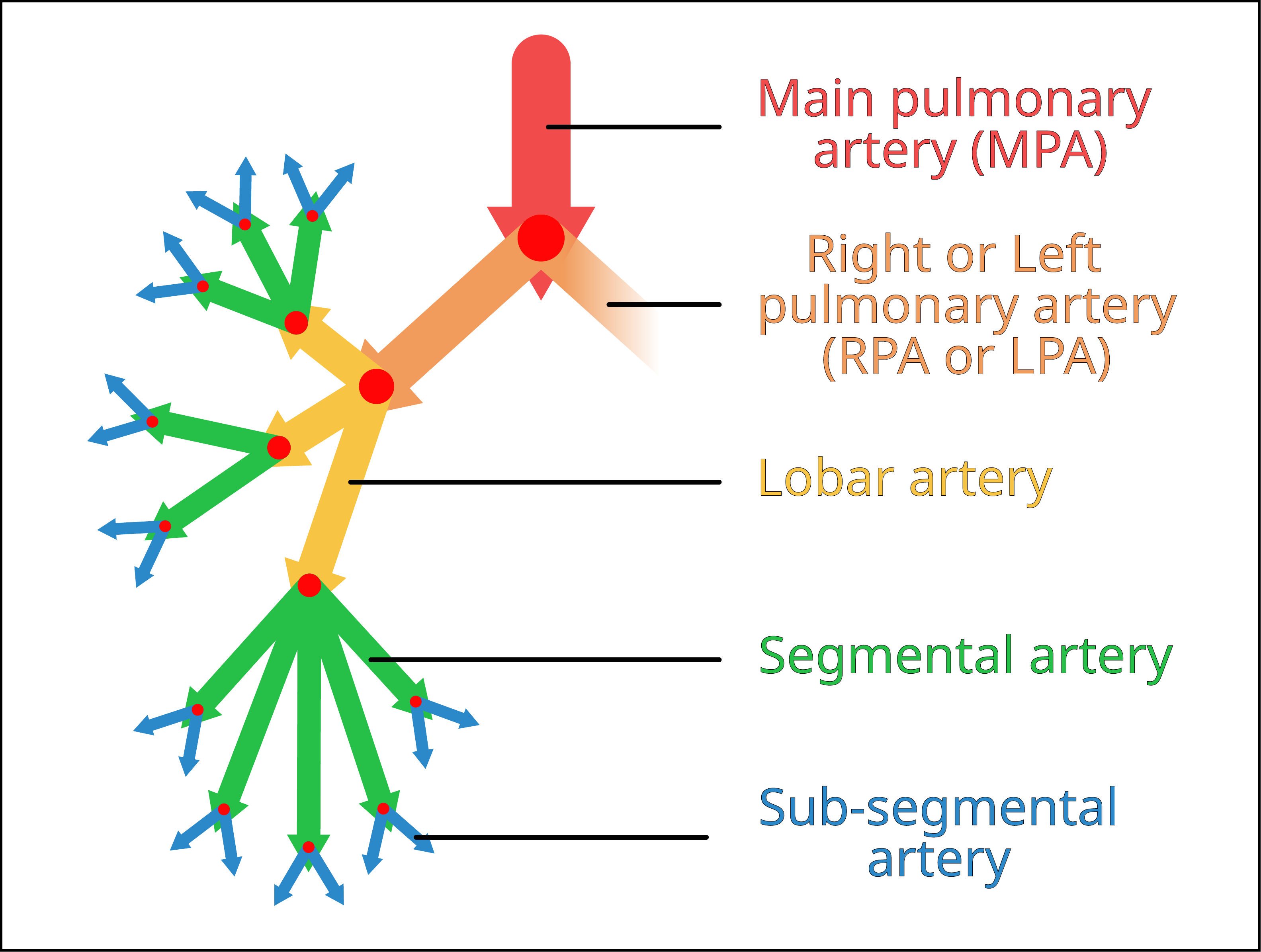}
\caption{Hierarchical levels of pulmonary arteries, each level is presented with a different color.}
\label{fig:hier_level}
\end{figure}

The hierarchical labeling process enriches the oriented vascular graph with anatomical information by assigning to each edge its laterality (right or left lung), lobar affiliation (upper, middle, or lower lobe), and hierarchical level within the PAT (e.g., main, Right of left, lobar, segmental, subsegmental etc) (see Fig.\ \ref{fig:hier_level}).
The root of the graph is defined as the MPA. Laterality is determined by intersecting each edge with the right and left lung masks. The first bifurcation distal to (after) the MPA defines the right and left pulmonary arteries (RPA and LPA, hierarchical level 2). Lobar affiliation is assigned by intersecting terminal arteries with the lobe masks and propagating the corresponding label upstream along the parent edges until the right or left pulmonary artery is reached (hierarchical level 3). Edges receiving multiple lobe labels are classified interlobar and will remain level 3. Hierarchical levels within each lobe are then refined using morphological criteria to distinguish trunks from bifurcations. Trunk arteries inherit the hierarchical level of their parent artery, whereas bifurcations are assigned the parent level + 1.

Using the artery diameter ($d_\text{artery}$), the diameter length ($L_\text{artery}$) and the parent artery diameter ($d_\text{parent}$), two morphological ratios are calculated to identify trunk structure, the diameter ratio :  
\begin{equation}\label{equa:diameter}
r_\text{diam} = \frac{d_\text{artery}}{d_\text{parent}},
\end{equation}
and the length ratio:
\begin{equation}\label{equa:length}
r_\text{len} = \frac{L_\text{artery}}{d_\text{artery}}.
\end{equation}
An artery is classified as a trunk if $r_\text{diam} \geq \theta_d$ and $r_\text{len} \leq \theta_l$. We evaluated \(\theta_d \in \{0.5, 0.6, 0.7\}\) and \(\theta_l \in \{1.5, 2.0, 2.5\}\). The optimal parameter pair was selected by minimizing the difference between the mean number of detected segmental arteries per lobe and the expected anatomical number reported in the literature \cite{michaud2023}. The selected thresholds are \(\theta_d = 0.7\) and \(\theta_l = 1.5\). Detailed quantitative results are reported in Section \ref{subsec:labeling}.

\subsubsection{PE characterization}
The PE feature extraction process performs a quantitative analysis of thrombi by identifying each thrombus, calculating its volume, and positioning it relatively to the PAT labeling. For each edge, the \textbf{volumetric obstruction} ratio is obtained by quantifying the volume of embolism voxels intersecting with the volume of the corresponding artery voxel in the 3D masks. Additionally, the transversal obstruction is quantified using Curved Planar Reformation (CPR) based cross-sectional measurements along the edge’s centerline, capturing localized luminal narrowing that volumetric metrics don't depict. The 3D artery segment is computationally "straightened" along its centerline using CPR, generating a stack of perpendicular cross-sections. For each CPR slice perpendicular to the centerline, the transversal obstruction is the clot area divided by the total area of the artery. The maximum transversal obstruction of the artery is an important feature to characterize the impact of the obstruction inside each edge since it represents the most severe cross-sectional narrowing anywhere along the artery, regardless of total clot volume, and resulting volumetric obstruction. 

\subsection{Biomarkers extraction}\label{subsec:biomarkers_extraction}

The biomarkers extracted by the proposed pipeline are organized into two categories: \textit{local biomarkers}, defined at the edge level of the PAT graph, and \textit{patient-level biomarkers}, expressed as global percentages or literature scores and computed from the local measurements.

\subsubsection{Local Biomarkers}
The first three stages of the pipeline (pulmonary artery graph modeling, hierarchical labeling and the PE characterization) provide a complete set of local biomarkers, grouped into three categories. 
% artery features
\textbf{Arteries features} describe the morphological properties of each artery, including its length and diameter statistics (mean, minimum, and maximum).
% Location features
\textbf{Location features} describe the position of the artery within the hierarchical structure of the PAT (Fig.\ \ref{fig:hier_level}) and include:
\begin{itemize}
    \item Laterality (right or left lung);
    \item Lobar affiliation (right upper lobe (RUL), right middle lobe (RML), right lower lobe (RLL), left upper lobe (LUL), left lower lobe (LLL) or None for the main pulmonary artery and the two primary branches);
    \item Hierarchical level, from 1 (MPA), to distal subsegmental arteries (level $\geq$ 5).
\end{itemize}
% PE features 
\textbf{PE features} quantify thrombotic obstruction at the edge level and include the clot volume, the volumetric obstruction ratio, and the maximum transversal obstruction, as defined in the previous section.

\subsubsection{Patient-level Biomarkers}
From the local biomarkers, the following patient-level indicators are computed:
\begin{itemize}
    \item Qanadli and Mastora obstruction scores, calculated using the maximum transversal obstruction;
    \item Total embolic volume (TEV) distribution across pulmonary lobes, to assess the spatial heterogeneity of clot burden;
    \item TEV distribution across hierarchical levels, to characterize the proximal or distal predominance of embolic involvement.
\end{itemize}

The automatic calculation of the Qanadli and Mastora scores follows their original definitions but is not straightforward, due to the mismatch between the idealized anatomical structure assumed by these scores and the graph-based decomposition accessible through automatic reconstruction. 

Several distinctions encoded in the original formulations --such as hierarchical levels or a fixed theoretical number of segments-- cannot be reliably identified in a systematic way. Rather than imposing these constraints a priori, we adopt a functional representation driven by the graph structure: levels are merged when necessary, and the number of segments considered is patient-specific, based solely on the elements effectively detected. This choice avoids heuristics dependent on fine anatomical details that are difficult to observe automatically, while preserving the overall logic of the reference scoring systems.

For both scores, the degree of obstruction is defined using the automatically measured maximum transversal obstruction. For Qanadli, a threshold-based scheme is applied: values below 25\% are considered not obstructed, values between 25\% and 75\% correspond to partial obstruction (weight of 1) and values above 75\% correspond to total obstruction (weight of 2). For Mastora, obstruction grades strictly follow the original definition: 1 (0–24\%), 2 (25–49\%), 3 (50–74\%), 4 (75–99\%), 5: (100\%). Finally, to assess the impact of these choices and the relevance of the automatic scores, an expert manually calculated the scores for ten patients. Manual and automatic scores were then compared using Bland-Altman analysis and Spearman correlation, allowing the evaluation of both agreement and strength of association between the two approaches.

TEV repartition by lobes and hierarchical levels provides a more detailed spatial characterization of embolic burden. This spatial information can help distinguish between proximal and distal PE patterns, which have different prognostic implications \cite{paez-carpio2025}.

\section{Dataset}\label{sec:dataset}
The study relies on the PERSEVERE dataset, which includes 431 patients diagnosed with acute pulmonary embolism at the emergency department of CHU Saint-Étienne between April 2014 and June 2020. All patients were diagnosed by CTPA and admitted within 72 hours to the Vascular and Therapeutic Medicine unit. The dataset also provides extensive clinical information, including pulmonary embolism severity level, sPESI score \cite{Jimenez2010}, as well as troponin and BNP measurements.  The final cohort received approval from the local ethics committee (IRBN262017/CHUSTE).

\subsection{Image Acquisition and Reconstruction Protocol}
CTPA data were reconstructed using soft convolution kernels (e.g., B30f, B40f) with a mediastinal window setting. Typical acquisition parameters included a tube voltage of 120 kVp and tube current ranging 102-661 mAs, with a slice thickness of 1 mm. An iodinated contrast agent (Xenetix) was administered at the arterial phase.

\subsection{Exclusion criteria}
Several exclusion criteria were applied to ensure image quality and data consistency. To maintain homogeneity of acquisition protocols, only patients scanned on Siemens scanners of model \textit{Somatom Definition} or \textit{Somatom Definition AS} were retained, given the limited number of examinations performed on other devices (15 patients; see Appendix I%\ref{annexe:dataset} TO DO put back reference for final submission
). Among the remaining patients ($n=416$), 24 were excluded due to unusable scans, including truncated heart or lungs, absence of visible emboli, missing arterial phase, or severe imaging artifacts. In addition, 39 patients without available troponin measurements were excluded, as this biomarker is required for downstream analyses investigating correlation between imaging-derived biomarkers and established indicators of cardiac stress. After applying all exclusion criteria, 353 patients were retained for the final PERSEVERE cohort/ %(see Annexe Fig.\ \ref{fig:arbre_decisions_353}). 

\subsection{CTPA preprocessing}
All CTPA volumes were converted from DICOM to NIfTI format and resampled to the median voxel spacing of the dataset (0.72 × 0.72 × 0.5 mm). Volumes were then cropped with a margin around the lung masks generated with TotalSegmentator, to reduce memory usage and restrict the field of view to the relevant thoracic region.

\subsection{Ground-truth annotations and segmentation}
Several types of masks were required to apply the proposed pipeline (see Fig.\ \ref{fig:general_pipeline} A), including lungs, lobes and pulmonary arteries and emboli. Lung and lobe masks were obtained automatically, whereas arterial and embolic structures required dedicated annotations for supervised learning and validation. The generation of these annotations and the resulting segmentations are detailed below.

\begin{table}[htbp!]
\centering
\caption{Counts of annotations and segmentations per structure, including Betti numbers statistics}
\label{tab:dataset}
\begin{tabular}{l c c}
\toprule
\textbf{Structure} & \textbf{Annotations} & \textbf{Segmentations} \\
\midrule
Arteries & 30 & 353 \\
\quad \textit{Betti numbers} & & \\
\quad \quad $\beta0$ (connected components) & 1.0 $\pm$ 0.0 & 1.0 $\pm$ 0.0 \\ % after correction for segmentation 1.0 $\pm$ 0.0
\quad \quad $\beta1$ (cycles or tunnels) & 1.2 $\pm$ 1.5 & 0.7 $\pm$ 1.2 \\ % after correction for segmentation 0.5 $\pm$ 0.9
\quad \quad $\beta2$ (holes) & 0.0 $\pm$ 0.0 & 0.0 $\pm$ 0.0 \\ % after correction for segmentation 0.0 $\pm$ 0.0 
\quad \quad Patients with $\beta1 >$ 0 & 16 (53\%) & 138 (39\%) \\ % after correction for segmentation 111 (31.4\%)
\midrule
Embolies & 24 & 353 \\
\midrule
Lungs \& lobes & - & 353 \\
\bottomrule
\end{tabular}
\end{table}

\subsubsection{Manual annotations and iterative refinement}\label{dataset:annotation}
A first subset of 10 patients was manually annotated using 3D Slicer for pulmonary emboli and a 3D Slicer semi-automatic plugin for pulmonary arteries \cite{desligneris2025}. These annotations were used to train initial nnU-Net models. Based on the first iteration, arterial masks from 20 additional patients were corrected, and emboli masks from 14 patients were manually refined. After this second and final iteration, a total of 30 annotated arterial trees and 24 annotated embolic volumes were obtained. 

\subsubsection{Cohort-wide segmentation}
The best-performing nnU-Net models from the iterative process were then applied to the entire cohort of 353 patients.
\begin{itemize}
    \item Lung and lobe masks were generated automatically for all patients using TotalSegmentator \cite{wasserthal2023}.  
    \item Pulmonary arteries were segmented using nnU-Net Model 1bis trained on 30 arterial annotations.  
    \item Pulmonary emboli were segmented using nnU-Net Model 4 trained on 24 annotated patients with combined embolus, artery, and lung masks.  
\end{itemize}
Details regarding training, selection and validation are provided in  Appendix II%\ref{annexe:segmentation} TO DO put back reference for final submission
, with segmentation results reported in Appendix III%\ref{annexe:results-segmentation} TO DO put back reference for final submission
.

\subsubsection{Segmentation post-processing}
All segmentation masks were post-processed to enforce anatomical and topological consistency prior to graph construction and analysis. For pulmonary arteries, only the largest connected component was retained in order to ensure a single connected arterial tree ($\beta_0 = 1$). Small isolated components were discarded. For pulmonary emboli, voxels located outside the arterial mask were removed, ensuring that emboli were strictly embedded within the arterial lumen. Topological integrity of the arterial masks was further assessed using Betti numbers. After post-processing, all arterial segmentations exhibited no internal cavities ($\beta_2 = 0$). A single voxel-sized cavity was detected in one patient and corrected manually. Cycles ($\beta_1 > 0$) were occasionally present in the arterial masks. In 48 patients, residual cycles remained and were manually corrected by an expert to allow successful execution of the pipeline, as described in Section~\ref{subsubsec:res_voreen}. A summary of the available annotations and cohort-wide segmentations for each anatomical structure, together with topological statistics derived from the final masks, is reported in Table~\ref{tab:dataset}.

\section{Results}\label{sec:result}

This section reports the evaluation of the proposed pipeline. Graph construction and artifact correction are first compared with the VesselGraph pipeline from the Voreen framework, and hierarchical labeling is then assessed with respect to the known pulmonary arterial anatomy. Finally, automatically computed Qanadli and Mastora scores are compared with manual experts annotations. 

\subsection{Graph modeling and artifact correction}\label{subsubsec:res_voreen}
As a baseline for graph construction, we used the VesselGraph pipeline \cite{drees2021}, available within the Voreen framework \cite{meyer-spradow2009}. This pipeline takes a binary vessel mask as input to extract the vascular centerline and produces a centerline-based graph representation. The comparison with VesselGraph was performed exclusively on a subset of 24 manually annotated cases, obtained after the first iteration of the annotation refinement process (see Section~\ref{dataset:annotation}). This restriction is due to the fact that VesselGraph graph generation requires manual interaction through the graphical interface, making large-scale processing impractical.
For the full PERSEVERE dataset (353 segmentations), only our internal graph construction and cleaning pipeline was applied.

\begin{table}[htbp!]
\centering
\caption{Construction and cleaning statistics}
\label{tab:vorren_vs_cleaning}
\begin{tabular}{c c c c c}
\toprule
\makecell{\textbf{Graph} \\ \textbf{type}} &
\textbf{Nodes} &
\textbf{Edges} &
\makecell{\textbf{Artifact} \\ \textbf{on MPA}} &
\textbf{Cycles} \\
\midrule
\multicolumn{5}{l}{\textbf{24 annotations}} \\
\midrule
VesselGraph& 370 ± 108 & 370 ± 109 & 4.0 ± 4.1 & 1.1 ± 1.4 \\
Constructed graph & 370 ± 130 & 370 ± 130 & 1.4 ± 1.7 & 1.2 ± 1.5 \\
Corrected graph & 292 ± 112 & 291 ± 112 & 0.0 ± 0.0 & 0.1 ± 0.3 \\
\midrule
\midrule
\multicolumn{5}{l}{\textbf{353 segmentations}} \\
\midrule
Constructed graph & 243 ± 66 & 243 ± 66 & / & 0.7 ± 1.1 \\
Corrected graph & 193 ± 57 & 192 ± 57 & / & 0.2 ± 0.6 \\
\bottomrule
\end{tabular}
\end{table}

Table \ref{tab:vorren_vs_cleaning} reports graph statistics for two experimental settings, including the number of nodes and edges, the number of spurious branches originating from the main pulmonary artery (MPA artifacts), and the number of cycles detected in the resulting graphs. 

On the 24 annotated cases, the number of nodes and edges is nearly identical between VesselGraph and our constructed graph before cleaning, indicating that the graph construction step does not alter graph topology. After cleaning, the number of nodes and edges is reduced by $22\% \pm 5\%$, corresponding to the removal of artifactual branches rather than anatomically relevant structures. This interpretation is supported by the complete elimination of MPA-related artifacts after cleaning, whereas VesselGraph leaves $4.0 \pm 4.1$ such artifacts on average. Cycle analysis shows comparable values before cleaning for VesselGraph and our constructed graphs (approximately one cycle per patient on average). After cleaning, the number of cycles is strongly reduced, with a mean of $0.1 \pm 0.3$ cycles per patient and a maximum of one remaining cycle.

For the full dataset, cycles were detected in 138 patients before cleaning and in 48 patients afterward. These remaining cases correspond to atypical anatomical configurations or severe pathological deformations that require manual correction, after which the pipeline could be successfully applied.

\begin{figure}[ht!]
\centering
\includegraphics[width=\columnwidth]{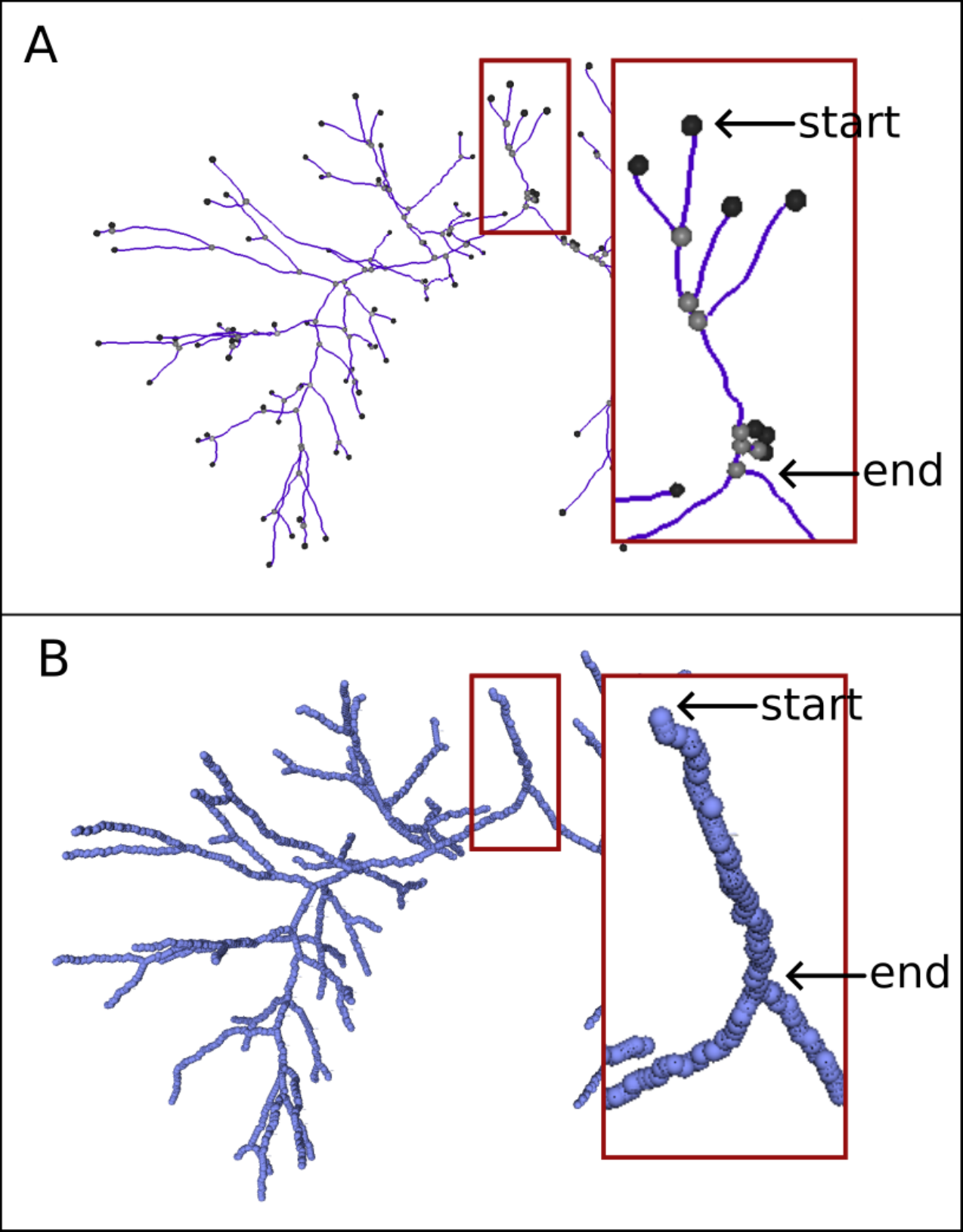}
\caption{Comparison of vascular graphs obtained on the same patient with a zoom view of the main pulmonary artery. (A) Graph generated by the Voreen VesselGraph pipeline, containing artifactual branches on the MPA. (B) Graph after our cleaning step, showing the removal of these artifacts.}
\label{fig:voreen}
\end{figure}
Figure \ref{fig:voreen} illustrates a qualitative comparison of patient 0028 from our annotated dataset. On the left, the graph produced by the Voreen VesselGraph pipeline is shown. On the right, the graph obtained after applying our cleaning step to the same initial graph is displayed. The MPA is highlighted in both representations. Several artifactual branches are visible on the MPA in the Voreen graph, whereas the graph after the cleaning step shows an artifact-free MPA. 

\subsection{Hierarchical labeling}\label{subsec:labeling}

\begin{table}[htbp!]
\centering
\caption{Mean number of segmental arteries}
\label{tab:number_segments}
\begin{tabular}{c c c c}
\toprule
\makecell[c]{\textbf{Lobe}} &
\makecell[c]{\textbf{Expected number of} \\ \textbf{segmental arteries}} &
\multicolumn{2}{c}{\textbf{\makecell{Mean number of segmental arteries\\automatically detected}}} \\
\cmidrule(lr){3-4}
& & \textbf{24 annotations} & \textbf{353 segmentations} \\
\midrule
RUL & 3 & 4.2 ± 1.8 & 3.4 ± 1.5 \\
RML & 2 & 2.6 ± 2.5 & 1.7 ± 1.0 \\
RLL  & 5 & 5.3 ± 1.8 & 4.9 ± 2.2 \\
LUL   & 5 & 5.2 ± 1.5 & 4.9 ± 1.2 \\
LLL   & 4 & 5.5 ± 2.1 & 5.3 ± 2.2 \\
\midrule
Total & 19 & 22.7 ± 4.9 & 20.2 ± 4.3 \\
\bottomrule
\end{tabular}
\begin{tablenotes}[flushleft]
\setlength\labelsep{0pt}
\footnotesize
\item RUL: Right Upper Lobe; RML: Right Middle lobe; RLL: Right Lower Lobe; LUL: Left Upper Lobe; LLL: Left Lower Lobe
\end{tablenotes}
\end{table}

Table \ref{tab:number_segments} reports the mean number of segmental arteries automatically detected per pulmonary lobe, compared with expected anatomical values. Results are presented separately for the subset of 24 manually annotated cases and for the full dataset of 353 segmentations.

For the 24 annotated cases, the mean number of detected segmental arteries is close to the expected anatomical configuration for all lobes, with deviations generally below one segment per lobe. A slight over-segmentation tendency is observed, particularly in the right upper and left lower lobes.

For the full dataset, detected segment counts are overall closer to the expected anatomical values, with reduced deviations for most lobes. This difference likely reflects the composition of the datasets: the annotated subset was selected to include anatomically complex and pathological cases, whereas the full cohort contains a larger proportion of simpler anatomies.

It should be noted that the expected number of segmental arteries corresponds to a simplified anatomical reference derived from the literature and assumes a healthy pulmonary arterial tree. In practice, pathology-related alterations, such as artery dilatation upstream blood clot, or an unusual branching pattern, lead to deviations from the anatomical model \cite{michaud2023}. The method parameters were selected to favor the closest anatomically plausible configuration rather than enforcing a fixed number of segments (see Section \ref{subsec:hier_labeling}).

\subsection{Literature scores}

\begin{table}[ht]
\centering
\caption{Bland-Altman and Spearman correlation}
\label{tab:scores}
\begin{tabular}{c c c c c}
\toprule
\textbf{Scores} &
\makecell{\textbf{Mean diff}\\\textbf{$\pm$ SD}} &
\makecell{\textbf{Lower}\\\textbf{LoA}} &
\makecell{\textbf{Upper}\\\textbf{LoA}} &
\makecell{\textbf{Spearman correlation}\\ \textbf{$\rho$ (p-value)}} \\
\midrule
Qanadli & $4.70 \pm 10.45$ & -15.78 & 25.18 & 0.83 (0.0029)\\ 
\midrule
Mastora & $-2.45 \pm 7.39$ & -16.93 & 12.02 & 0.86 (0.0015)\\ 
\bottomrule
\end{tabular}
\end{table}

Table \ref{tab:scores} reports the results of the Bland-Altman analysis and Spearman correlation between manual and automatic scoring methods. 
% Bland Altman
The Bland-Altman analysis indicates better agreement for the Mastora score than for the Qanadli score. The Mastora score showed smaller systematic bias (mean difference: -2.45\% vs 4.70\%) and narrower limits of agreement (range: 28.95\% vs 40.96\%), reflecting more consistent agreement between manual and automatic measurements. The mean differences indicate a slight underestimation for the Mastora score and a slight overestimation for the Qanadli score.
% Spearman
Spearman correlation coefficients were used to assess the monotonic association between manual and automatic scores. Strong correlations were observed for both scoring systems, with $\rho = 0.83$ ($p = 0.0029$) for the Qanadli score and $\rho = 0.86$ ($p = 0.0015$) for the Mastora score, indicating statistically significant agreement the two measurement approaches.

\subsection{Patient level biomarkers}

\begin{figure}[ht!]
\centering
\includegraphics[width=\columnwidth]{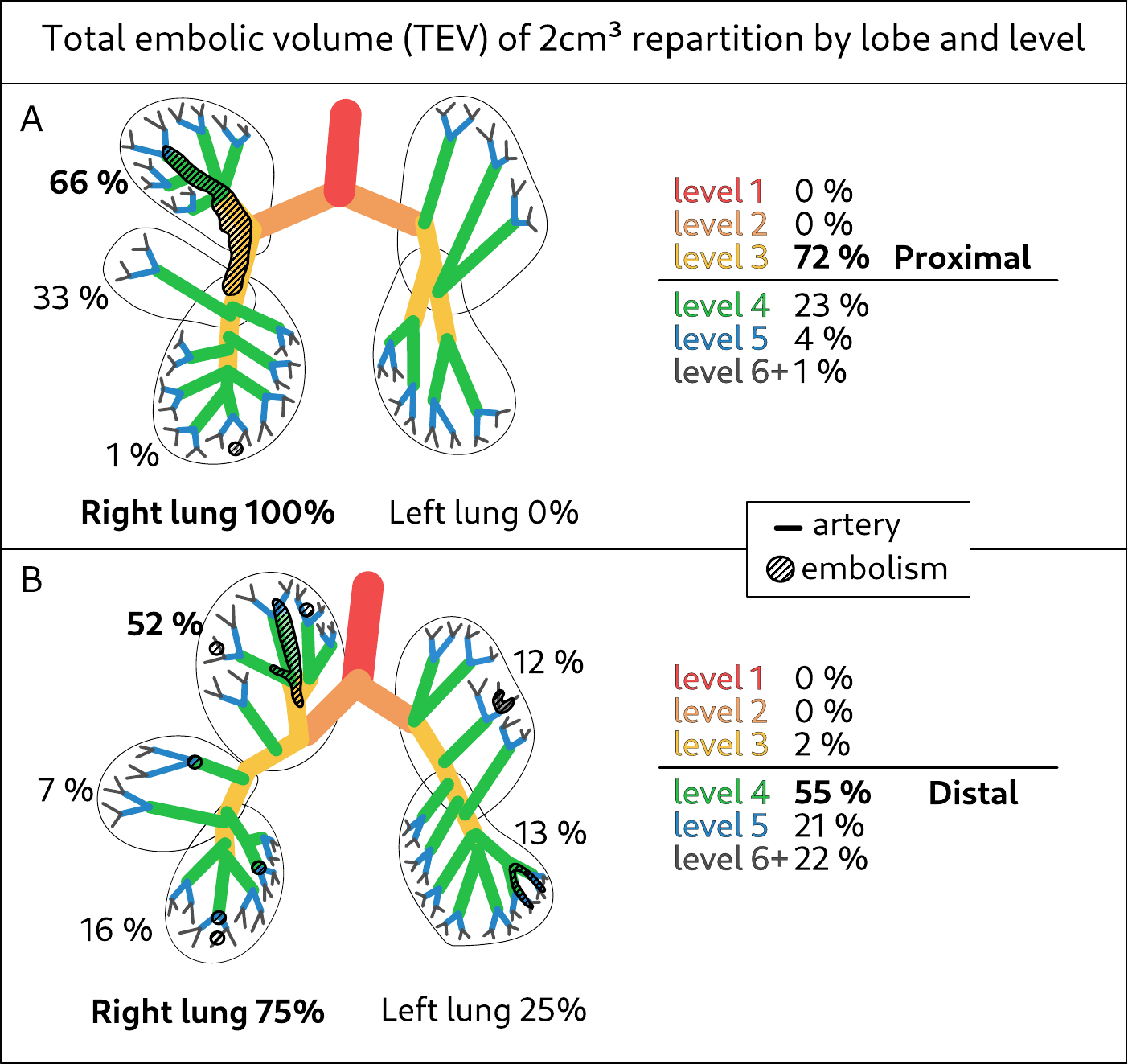}
\caption{Example case of global biomarkers extracted to characterize the total embolic volume repartition by lobes and levels.} 
\label{fig:global_biomarkers}
\end{figure}

Figure \ref{fig:global_biomarkers} illustrates two patients with identical total embolic volume (TEV$= 2cm^3$), but distinct spatial distributions of pulmonary emboli across lobes and hierarchical arterial levels. Emboli located before the segmental arteries (level 4) are considered proximal, whereas emboli beyond this level are considered distal. 

Although both patients present a predominance of embolic burden in the right upper lobe, patient A exhibits mainly lobar-level emboli, while patient B shows predominantly segmental-level involvement. These differences are revealed through the local biomarkers extracted by the proposed pipeline, which provide a lobe- and level-resolved characterization of thrombotic load. As a result, the pipeline enables discrimination between proximal and distal embolic patterns that are not captured by total embolic volume alone.

\subsection{Execution time}

\begin{table}[htbp!]
\centering
\caption{Execution time on the PERSEVERE dataset}
\label{tab:execution_time}
\begin{tabular}{l c}
\toprule
\textbf{Step} & \textbf{Execution time} \\
\midrule
\multicolumn{2}{l}{Pulmonary Artery Graph Modeling (1)} \\
\quad Preprocessing (a) & 15.5s ± 4.0s \\
\quad Graph construction (b) & 16.9s ± 4.4s \\
\quad Graph correction (c) & 17.3s ± 5.3s \\
\quad Graph orientation (d) & 12.1s ± 3.3s \\
\addlinespace[0.5em]
Total (a–d) & \makecell{1min01s ± 0min15s\\(61.8s ± 15.8s)} \\
\midrule
Hierarchical labeling (2) & \makecell{3min58s ± 1min32s\\(238.4s ± 92.7s)}\\
\midrule
PE characterization (3) & \makecell{3min48s ± 2min58s\\(228.8s ± 178.5s)}\\
\midrule
Patient-level feature extraction (4) & 0.0s ± 0.2s\\
\midrule
\textbf{Total Processing Time (1-4)}& \makecell{8min48s ± 3min21s\\(528.4s ± 201.9s)}\\
\bottomrule
\end{tabular}
\end{table}

On average for one patient, the execution time is approximately 1 minute for the pulmonary artery graph modeling, 4 minutes for the hierarchical labeling, and 4 minutes for the PE characterization. The time to compute the patient-level features is negligible once the enriched directed graph with all local features is completed. 

\section{Discussion}
This study presents an end-to-end pipeline for automatic extraction of imaging-based biomarkers from CTPA scans to characterize pulmonary embolism severity. By combining segmentation, graph-based vascular modeling, and hierarchical feature extraction, the proposed approach enables the computation of both local and patient-level biomarkers, including artery morphology, PE localization, and obstruction metrics, directly from routine imaging data.

A key contribution of this work lies in the construction of anatomically consistent pulmonary arterial graphs. Iterative nnU-Net training, combined with post-processing and expert validation, produced arterial and embolic segmentations with enforced connectivity and no internal cavities, enabling consistent graph construction and feature extraction. Compared with the Voreen VesselGraph pipeline, the proposed graph correction and hierarchical labeling steps effectively reduce artifactual branches and cycles while preserving anatomically plausible vascular structures. The pipeline enables the automatic computation of established PE severity scores (e.g., Qanadli, Mastora), which are rarely used in clinical practice due to their time-consuming manual calculation.
In addition, the extraction of TEV and its distribution across pulmonary lobes and hierarchical arterial levels provides complementary information that may better reflect the spatial extent and functional impact of the disease.
% perspective
Future work will include a dedicated clinical study to evaluate the correlation between automated imaged-based biomarkers and blood biomarkers (BNP, troponin), as well as their predictive value for patient outcomes.

Several limitations should be acknowledged. 
First, the pipeline relies on high-quality arterial segmentations. If the arterial mask does not form a single connected component or fails to correctly represent the vascular tree down to the segmental and subsegmental levels, the resulting directed PAT graph and derived hierarchical features may not be reliable. 
Second, the hierarchical labeling procedure depends on the accuracy of lung and lobe segmentations. Ambiguous terminal points, particularly near lobar boundaries or outside anatomical masks, remain challenging to resolve automatically. In addition, the balance between anatomical sensitivity and algorithmic robustness is difficult to achieve when applying morphology-based rules to distinguish arterial levels. 
Parameters were calibrated on a subset of 24 patients and may not fully capture the wide range of anatomical variations, especially in cases with severe pathological deformation of the arteries. This limitation particularly affects the distinction between level 3 and level 4 branches, where anatomical variability is well documented \cite{michaud2023}.
Finally, while agreement between automated and manual severity scores was demonstrated, this validation was performed on a limited sample of patients. Larger-scale studies will be required to fully assess robustness and generalizability across diverse patient populations and acquisition protocols.

\section{Future work}
Several directions for improvement and clinical translation can be identified. In the short term, improving the robustness and generalizability of the segmentation models is an important issue. Incorporating more diverse training data, including cases with severely damaged lungs and atypical vascular anatomies, is expected to reduce segmentation errors and the occurrence of anatomically incorrect cycles. This would directly limit the need for expert intervention on artery mask and improve the reliability of downstream graph construction.

The hierarchical labeling procedure could also be refined. Future work could explore graph neural networks (GNNs) to classify nodes (hierarchical levels) by leveraging both neighborhood relationship and arteries features (morphological properties). Given the newly acquired labeled graph dataset, this new approach could replace the current heuristic rules, improving adaptability to anatomical variability and pathology-induced vascular deformation. Increasing the robustness of hierarchical labeling would directly enhance proximal versus distal characterization of embolic burden.

From a translational perspective, the development of software enabling interactive visualization of the final arterial graph together with associated local and patient-level biomarkers would allow clinicians to directly relate quantitative measurements to the underlying vascular structure. Such visualization, as preliminarily illustrated in Fig.~\ref{fig:global_biomarkers}, could support quality control, facilitate clinical interpretation, and increase trust in automated severity assessment.

Finally, the extracted biomarkers provide a foundation for developing and validating novel imaging-based risk stratification models. A dedicated clinical study will be conducted to assess their added value, particularly in intermediate-risk patients, where therapeutic decision-making remains challenging. By integrating both local and global vascular features, such models may improve outcome prediction compared with current scoring systems. In the longer term, this approach could support more personalized management strategies by identifying patients who may benefit from advanced interventions or closer monitoring based on individual embolic burden and vascular alterations.

\section*{Code availability}
The pipeline repository will be available at \href{https://gitlab.in2p3.fr/MDL/pulmonarytreegraph}{https://gitlab.in2p3.fr/MDL/pulmonarytreegraph}.

\section* {Conflicts of interest}
The authors declare no conflicts of interest related to this study.

\section*{References}
\bibliographystyle{IEEEtran}
\bibliography{manuscript}

\end{document}

\begin{appendices}

\section{Dataset}

Table~\ref{tab:manufacturers} summarizes the distribution of scanner manufacturers and models used for CTPA acquisition. To limit acquisition-related variability, only examinations acquired on Siemens scanners were included in the final cohort. In particular, scans acquired on the \textit{Somatom Definition} and \textit{Somatom Definition AS} models were retained.

\begin{table}[htbp!]
\centering
\caption{Distribution of patients per manufacturer and scanner model.}
\label{tab:manufacturers}
\setlength{\tabcolsep}{3pt}
\begin{tabular}{c c c}
\toprule
\textbf{Manufacturer} & \textbf{Model} & \textbf{\makecell{Number of\\patients}} \\
\midrule
Philips & Brilliance 40 & 1 \\
\midrule
\multirow{4}{*}{\makecell{GE}}  & BrightSpeed QX/i & 1 \\
 & Optima CT540 & 1 \\
 & Optima CT660 & 2 \\
 & Revolution EVO & 3 \\
\midrule
\multirow{3}{*}{Siemens} & Sensation 16 & 7 \\
 & Somatom Definition & 29 \\
 & Somatom Definition AS & 387 \\
\bottomrule
\end{tabular}
\end{table}

\section{Development and training of segmentation models}
\label{annexe:segmentation}
\begin{figure}[htbp!]
\label{fig:training_protocole}
\centering
\includegraphics[width=\columnwidth]{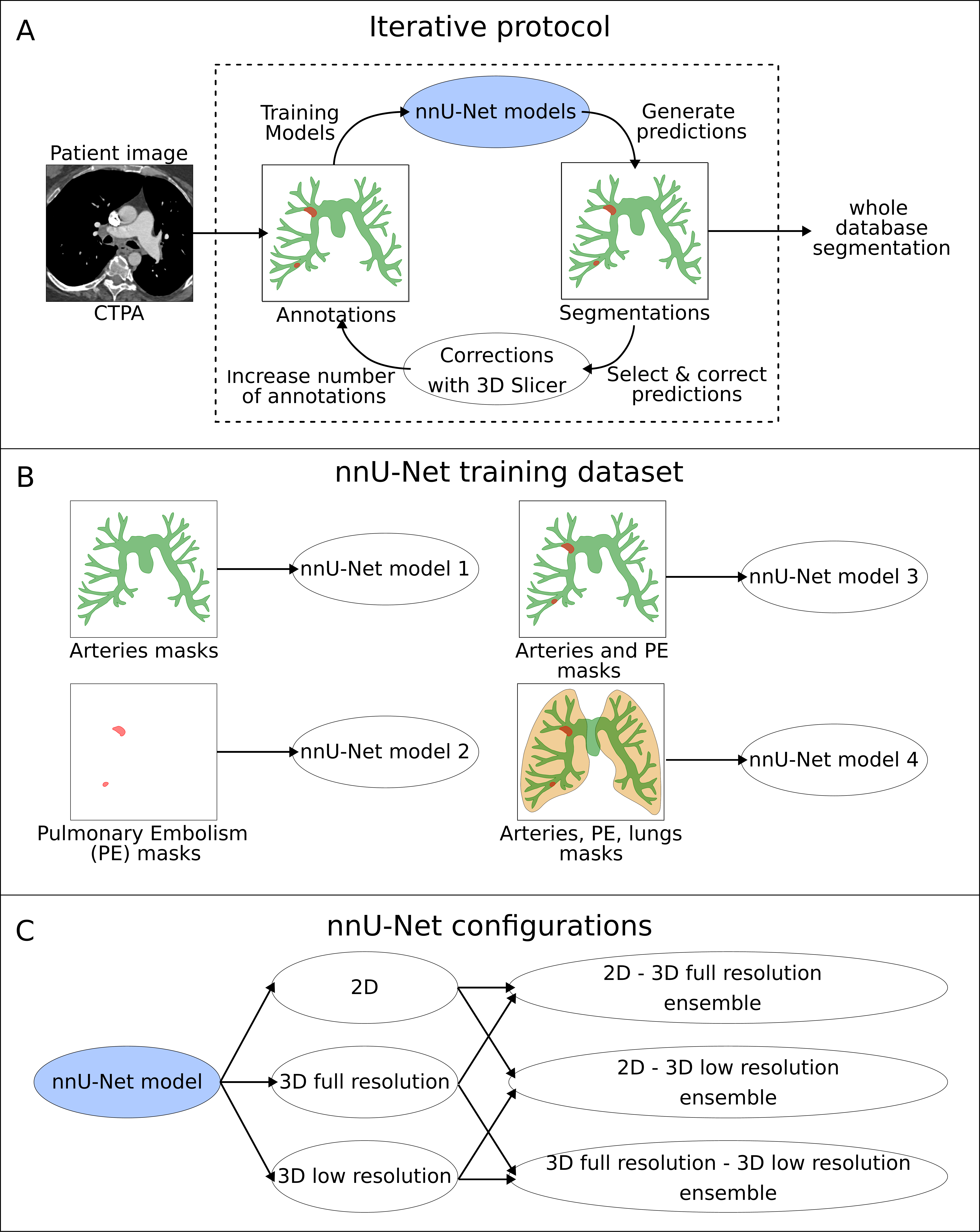}
\caption{Iterative training and configuration of nnU-Net segmentation models. 
(A) Iterative annotation and training protocol: a subset of patients is annotated, used to train nnU-Net models, predictions are reviewed and corrected, and the number of annotations is progressively increased until a final model is selected to segment the full database. 
(B) Anatomical contexts used to train the different nnU-Net models, based on combinations of artery, pulmonary embolism (PE), and lung masks. 
C) nnU-Net training configurations evaluated for each model: 2D, 3D low-resolution, 3D full-resolution, and their corresponding ensembles.}
\end{figure}

The development of segmentation models followed an iterative process summarized in Figure \ref{fig:training_protocole} A. A subset of patients was annotated, models were trained, segmentation quality was assessed, and predictions were corrected to progressively increase the number of annotations. By iteratively expanding the annotated dataset, we aimed to obtain a good quality model to be applied later to the full dataset. Models were trained using nnU-Net’s built-in pipeline, which includes automatic preprocessing steps such as resampling, normalization, and patch-based training. Manual corrections were performed either by direct editing or using the VSMOD 3D Slicer plugin \cite{desligneris2025}.

Four nnU-Net models were trained at each iteration, differing in the anatomical context provided during training (Fig.~\ref{fig:training_protocole}B):
\begin{itemize}
    \item Model number 1 : trained only on artery masks (label 1).
    \item Model number 2 : trained only on PE masks (label 1).
    \item Model number 3 : trained on combined artery and PE masks (label 1 for arteries, and label 2 for PE).
    \item Model number 4 : trained on artery, PE and lungs masks (label 1 for arteries, label 2 for PE and label 3 for lungs).
\end{itemize}

For each model, nnU-Net automatically generated three architectures: 2D, 3D low-resolution, and 3D full-resolution. All trainings were performed with 5-fold cross-validation to ensure robust estimation of generalization performance. In addition, ensemble predictions were generated for each pair of configurations (see Fig.\ \ref{fig:training_protocole}). To determine the best configuration, nnU-Net computed the mean foreground Dice on the validation results, without post-processing, and selected the configuration with the highest value (see Section \ref{subsubsec:res-best-config}). The mean foreground Dice corresponds to the average Dice across all non-background classes.

Once the best nnU-Net configuration was identified, segmentation of the arteries and PE were generated of the test datasets, and the mean Dice score was computed against the ground truth, after post-processing. To refine raw segmentation outputs, simple criteria were applied: for arteries, only the largest connected component was retained; for emboli, false positives located outside the arterial tree were removed.

Two iterations of model training were conducted:  
\begin{itemize}
    \item Iteration 1: At this stage, manual annotations were available for 10 patients. The four nnU-Net models were trained using different combinations of artery, embolus and lungs masks from 8 patients, while 2 patients were kept for testing. This exploratory phase aimed to establish baseline performance and identify promising configurations.  
    \item Iteration 2: With an expanded dataset of 30 annotated arterial trees and 24 embolus masks, five nnU-Net models were trained (Models 1 to 4 and Model 1bis), with varying inclusion of arteries, emboli, and lung masks. Models 1, 2, 3, and 4 were trained on 19 of the 24 patients with combined arteries and embolus annotations, while the 5 remaining patients were kept for testing. Model 1bis was trained with 24 of the 30 artery annotations; with 6 patients reserved for testing. This constituted the main training phase, and the best resulting models were used for large-scale automatic segmentation.  
\end{itemize}
Results from both iterations are reported in Section \ref{annexe:results-segmentation}. 

\section{Segmentation results}
\label{annexe:results-segmentation}

\subsection{Best configuration}\label{subsubsec:res-best-config}

For each model, nnU-Net automatically selected the best-performing configuration among 2D, 3D low-resolution, 3D full-resolution, and their corresponding ensembles, based on the mean foreground Dice score computed on the validation set (Table \ref{tab:config_dice_foreground}). Dice scores were computed without post-processing, following the standard nnU-Net configuration selection procedure.

During the first training iteration, the 3D full-resolution configuration consistently achieved the best performance across all models. In the second iteration, the ensemble of 3D low-resolution and 3D full-resolution achieved the best performance for artery-only models (Models 1 and 1bis), while the 3D full-resolution configuration remained optimal for models trained with additional anatomical context.

\begin{table}[bp!]
\centering
\caption{Mean foreground Dice comparison between first and second 
training iterations for configuration selection, on validation 
results. Bold value indicates the best configuration for that 
iteration.}
\label{tab:config_dice_foreground}
\resizebox{\columnwidth}{!}{%
\begin{tabular}{c c c c c c c c}
\toprule
Iteration & Model & 2D & 3D low & 3D full & 
\makecell{ens. 2D\\3D low} & 
\makecell{ens. 2D\\3D full} & 
\makecell{ens. 3D\\low-full} \\
\midrule
\multirow{4}{*}{1st} 
 & 1    & 0.842 & 0.882 & \textbf{0.903} & 0.877 & 0.885 & 0.900 \\
 & 2    & 0.337 & 0.402 & \textbf{0.651} & 0.303 & 0.393 & 0.498 \\
 & 3    & 0.624 & 0.709 & \textbf{0.787} & 0.687 & 0.739 & 0.767 \\
 & 4    & 0.746 & 0.803 & \textbf{0.845} & 0.787 & 0.822 & 0.839 \\
\midrule
\multirow{5}{*}{2nd}
 & 1    & 0.893 & 0.932 & 0.940 & 0.922 & 0.924 & \textbf{0.943} \\
 & 1bis & 0.905 & 0.938 & 0.946 & 0.929 & 0.932 & \textbf{0.949} \\
 & 2    & 0.446 & 0.462 & \textbf{0.613} & 0.393 & 0.462 & 0.530 \\
 & 3    & 0.686 & 0.722 & \textbf{0.799} & 0.724 & 0.764 & 0.768 \\
 & 4    & 0.789 & 0.816 & \textbf{0.853} & 0.821 & 0.842 & 0.842 \\
\bottomrule
\end{tabular}%
}
\end{table}

\subsection{Best segmentation model}

Table \ref{tab:dice_on_test} reports the best Dice results obtained on the test datasets. During the first iteration, mean Dice values were computed on 2 test patients for both artery and embolus segmentation. During the second iteration, Dice scores were computed on 5 test patients for Models 1, 2, 3, and 4 and 6 test patients for Model 1bis. 
For artery segmentation, the highest Dice scores were obtained with \textbf{Model 1} during the first iteration, and with \textbf{Model 1bis} during the second iteration, corresponding to models trained only on artery annotations and using the largest number of annotated patients. For embolus segmentation, the best performance was achieved with \textbf{Model 4} during both iterations. Overall, for embolus segmentation, models trained with additional anatomical context, including artery and lung masks, consistently achieved higher performance across both iterations.

\begin{table}[htbp!]
\centering
\begin{threeparttable}
\caption{Performance of artery and embolus segmentation models: mean Dice score on test datasets. Bold values indicate the best-performing model per class for each iteration.}
\label{tab:dice_on_test}
\begin{tabular}{c c l c}
\toprule
Class & Iteration & Model & Mean $\pm$ SD \\
\midrule
\multirow{7}{*}{Arteries}
  & \multirow{3}{*}{1st}
    & \textbf{model\_1}   & \textbf{0.905 ± 0.047} \\
  &                             & model\_3   & 0.893 ± 0.061 \\
  &                             & model\_4   & 0.894 ± 0.062 \\
\cmidrule(lr){2-4}
  & \multirow{4}{*}{2nd}
    & model\_1   & 0.942 ± 0.043 \\
  &                             & \textbf{model\_1bis} & \textbf{0.943 ± 0.043} \\
  &                             & model\_3   & 0.934 ± 0.051 \\
  &                             & model\_4   & 0.935 ± 0.049 \\
\midrule
\multirow{6}{*}{Embolies}
  & \multirow{3}{*}{1st}
    & model\_2   & 0.595 ± 0.010 \\
  &                             & model\_3   & 0.640 ± 0.101 \\
  &                             & \textbf{model\_4}   & \textbf{0.668 ± 0.126} \\
\cmidrule(lr){2-4}
  & \multirow{3}{*}{2nd}
    & model\_2   & 0.778 ± 0.138 \\
  &                             & model\_3   &  0.810 ± 0.125 \\
  &                             & \textbf{model\_4}   & \textbf{0.817 ± 0.120} \\
\bottomrule
\end{tabular}
\begin{tablenotes}[flushleft]
\setlength\labelsep{0pt}
\footnotesize
\item Bold values indicate the best performing model for each training iteration. Model\_1bis was only trained in the second iteration. All values represent Dice from the segmentation with post-processing, against the dataset ground truth.
\end{tablenotes}
\end{threeparttable}
\end{table}

\end{appendices}

\section*{References}
\bibliographystyle{IEEEtran}
\bibliography{supplementary_material}